 \documentclass[pmlr,twocolumn,10pt]{jmlr} 

\usepackage{booktabs}
\usepackage{siunitx}
\usepackage{subcaption}
\usepackage[switch]{lineno}

\theorembodyfont{\upshape}
\theoremheaderfont{\scshape}
\theorempostheader{:}
\theoremsep{\newline}

\jmlrvolume{297}
\jmlryear{2025}
\jmlrworkshop{Machine Learning for Health (ML4H) 2025} 

 \title[Data Augmentation to Prevent MIA in TSF]{Embedding-Space Data Augmentation to Prevent Membership Inference Attacks in Clinical Time Series Forecasting}

\author{%
\Name{Marius Fracarolli} \Email{fracarolli@cl.uni-heidelberg.de}\\
\addr Department of Computational Linguistics, Heidelberg University, Germany\\
\Name{Michael Staniek} \Email{staniek@cl.uni-heidelberg.de}\\
\addr Department of Computational Linguistics, Heidelberg University, Germany\\
\Name{Stefan Riezler} \Email{riezler@cl.uni-heidelberg.de}\\
\addr Department of Computational Linguistics \\
\& Interdisciplinary Center for Scientific Computing (IWR), Heidelberg University, Germany
}

\begin{document}

\maketitle

\begin{abstract}
Balancing strong privacy guarantees with high predictive performance is critical for time series forecasting (TSF) tasks involving Electronic Health Records (EHR). In this study, we explore how data augmentation can mitigate Membership Inference Attacks (MIA) on TSF models. We show that retraining with synthetic data can substantially reduce the effectiveness of loss-based MIAs by reducing the attacker's true-positive to false-positive ratio. The key challenge is generating synthetic samples that closely resemble the original training data to confuse the attacker, while also introducing enough novelty to enhance the model’s ability to generalize to unseen data. We examine multiple augmentation strategies --- Zeroth-Order Optimization (ZOO), a variant of ZOO constrained by Principal Component Analysis (ZOO-PCA), and MixUp --- to strengthen model resilience without sacrificing accuracy.  Our experimental results show that ZOO-PCA yields the best reductions in TPR/FPR ratio for MIA attacks without sacrificing performance on test data. 
\end{abstract}

\paragraph*{Data and Code Availability}
We use public data from physionet.org (MIMIC-III, eICU). Our code is available at \href{https://github.com/MariusFracarolli/ML4H_2025_Data-Augmentation-to-Prevent-MIA-in-TSF/}{GitHub}\footnote{\url{https://github.com/MariusFracarolli/ML4H_2025_Data-Augmentation-to-Prevent-MIA-in-TSF/}}.

\paragraph*{Institutional Review Board (IRB)}
IRB approval is not required for public data.

\section{Introduction}

Membership Inference Attacks (MIAs) are a "de facto standard" attack scenario on the privacy of machine learning models \citep{CarliniETAL:22}. Such adversarial attempts to determine whether a given sample was part of the training set are a real-world problem in the application of machine learning to sensitive health data such as Electronic Health Records (EHRs). As is known since \cite{DinurNissim:03}, removing identifying information such as patients' names from a database is not enough to protect privacy, but instead random perturbations have to be applied to the outputs in order to protect privacy even in the simplest case of ``statistical'' queries such as averages over databases. If EHR data is used to train machine learning models, e.g., for medical time series forecasting (TSF), the randomization techniques provided in the framework of differential privacy \citep{Dwork:06,DworkRoth:14} allow giving strong guarantees on the information derivable from private training data when querying a machine learning algorithm. However, there is an unavoidable tradeoff between utility and privacy of trained machine learning algorithms \citep{JayaramanEvans:19,CaiETAL:21}. As shown by \cite{YeomETAL:18}, overfitting of machine learning models is sufficient to allow MIA under the simple assumption that the average training loss of the model, but no information about the training data is leaked. 

The goal of this work is to investigate the possibilities of data augmentation techniques to balance privacy protection against MIA and prediction utility for machine learning on EHR data, ideally with improvements on both criteria. 
As we will show empirically, although the simple loss-based MIA of \cite{YeomETAL:18} is weaker than attack scenarios based on shadow or reference models  \citep{ShokriETAL:17,YeETAL:22,CarliniETAL:22,ZarifadehETAL:24}, it is still effective for complex applications like multivariate TSF on sparsely sampled medical data.
Following \cite{CarliniETAL:22}, we start from an evaluation of the effectiveness of MIAs by considering the ratio of their true positive rate (TPR) over their false positive rate (FPR). The privacy of a trained machine learning model can be protected by decreasing the TPR/FPR ratio from two sides. First, the FPR rate can be increased by generating synthetic data and re-training the model on the augmented dataset. These data need to be sufficiently different from the training examples in order to allow a decrease in held-out loss of a re-trained model. Second, the TPR rate can be reduced by adding synthetic examples that are similar enough to the original training set such that fewer original training examples are correctly recognized as members by the attacker. The goal is to reduce the TPR/FPR ratio towards 1, where the attack becomes equivalent to random guessing.

The main contribution of our work is the presentation of an algorithm that optimizes a joint objective of privacy protection and utility by zeroth-order optimization (ZOO) in the space of neural embedding matrices of training examples. The goal of this algorithm is to guide data augmentation in the direction of synthesizing examples that lead to decrease of the TPR/FPR ratio. We exemplify our algorithm on the difficult regression problem of 24 hour TSF on multivariate clinical data. Our experimental comparison investigates the power of ZOO versus standard data augmentation algorithms like MixUp \citep{ZhangETAL:18mixup}. Best results are obtained by enhancing ZOO with information about the principal components in embedding space (ZOO-PCA). While MixUp achieves best performance on test data, the best reductions in TPR/FPR ratio for MIA attacks without sacrificing performance on test data are achieved by ZOO-PCA. Since ZOO-PCA directs the augmentation process along the most significant data variations, it also enjoys favorable convergence rates due to effectively performing sparse zeroth-order optimization \citep{BalasubramanianETAL:22}.

\section{Related Work}

Differential privacy has become a de-facto standard for theoretically well-founded privacy preservation in machine learning since it allows giving strong guarantees on the information derivable from private training data when querying a machine learning algorithm \citep{Dwork:06,Dwork:08,DworkRoth:14}. A shortcoming of randomization-infusion as done in differential privacy approaches is an unavoidable tradeoff between model utility and privacy protection. 
This privacy-utility gap has been proven to be unavoidable by \cite{CaiETAL:21}, and quantified empirically by \cite{AbadiETAL:16}. According to \cite{JayaramanEvans:19}, it leads to either models with limited accuracy loss and meaningless privacy guarantees, or to useless models with strong privacy guarantees. 

Data augmentation as alternative method for privacy protection has been discussed quite controversially. \cite{SablayrollesETAL:19} show that due to a reduction of the gap between training and held-out accuracy of models trained on augmented data, the effectiveness of loss-based attacks can be decreased. An even stronger claim is made by \cite{HintersdorfETAL:22} who show that using generative adversarial networks to synthesize data, a potentially infinite number of samples will be falsely classified as members of the training set by an attacker. However, \cite{KayaDumitras:21} claim that on complex tasks, data augmentation cannot provide a ``free lunch'' to defeat membership inference attacks. Besides different experimental settings, a conceptual difference between these approaches is the decision whether synthetic data should count as genuine members of the training set instead of as false positives \citep{YuETAL:21}. In our case, synthetic data are generated in the latent embedding space and do not even have correlates in the space of input representations. We thus believe synthetic embeddings generated by ZOO or MixUp to be fair and effective distractions of an attacker.

\section{Time Series Forecasting (TSF)} 

\paragraph{TSF with Transformers}

\begin{figure}[t!]
    \centering
    \includegraphics[width=0.74\columnwidth]{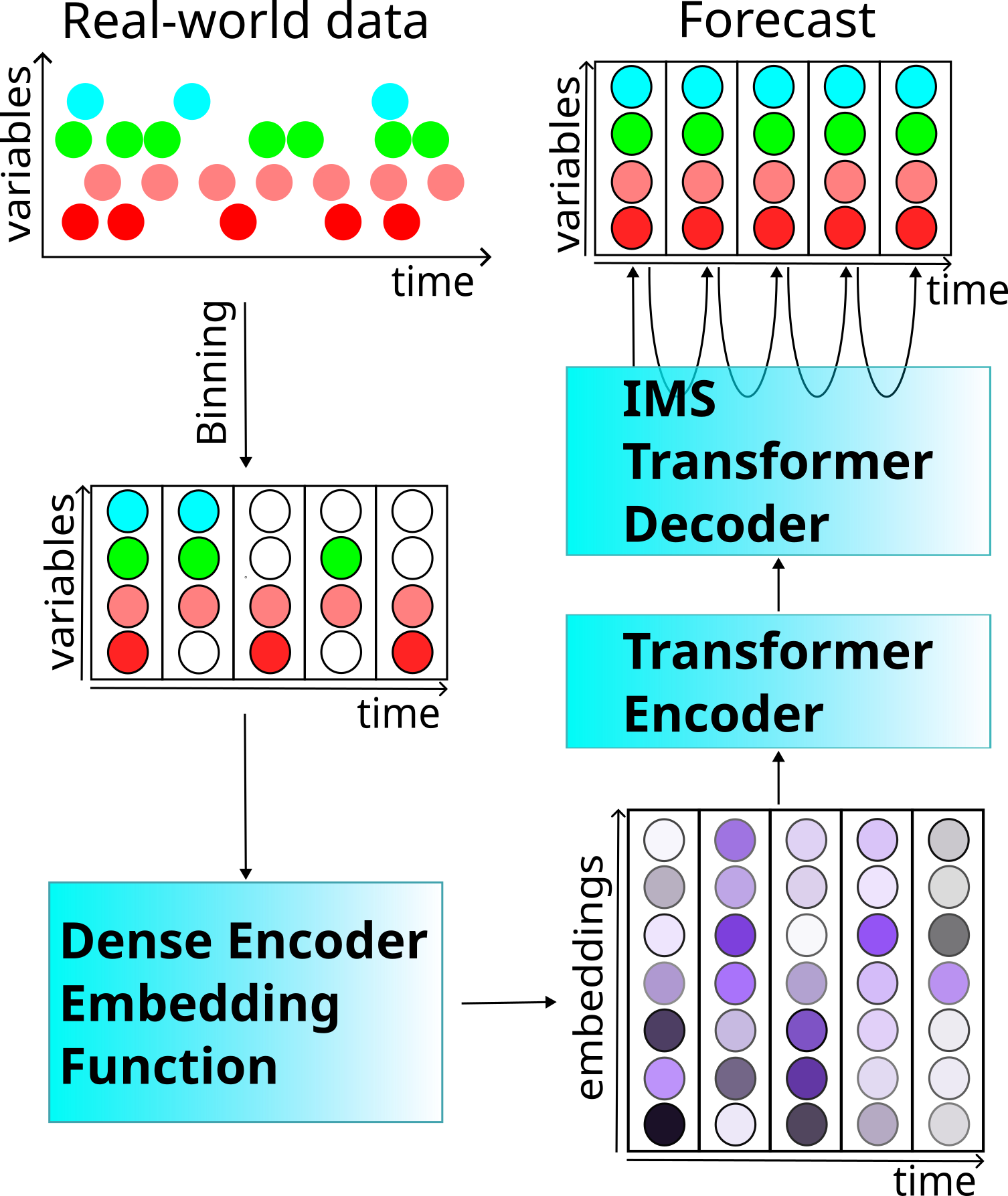}
    \caption{The process of time series forecasting for irregularly sampled medical time series data. Real-world data is binned into hourly buckets. The binned data is then transformed into embeddings using a Dense Encoder Embedding Function, capturing the embeddings relevant for the augmentation process. The embeddings are processed by a Transformer encoder, which learns contextual representations. An iterative multistep forecasting (IMS) decoder with autoregressive properties generates forecasts.}
    \label{fig:neural-architecture}
\end{figure}

To convert real-world data -- specifically, irregularly sampled medical time series -- into a long-term time series forecast (TSF), we use the implementation of \cite{StaniekETAL:24mlhc} which is based on a Transformer encoder-decoder architecture \citep{VaswaniETAL:17}. The architecture of this model is illustrated in Figure~\ref{fig:neural-architecture}.

In our dataset, each sample $S = \{(t_i, f_i, v_i)\}^n_{i=1}$ consists of triplets where:
\begin{itemize}
    \item  $t_i \in \mathbb R_{\geq 0}$ is a timestamp,
    \item $f_i$ belongs to a set of clinical variables $F$,
    \item $v_i \in \mathbb R$ is the corresponding measured value.
\end{itemize}

Thus, a data point $(t_\alpha , f_\alpha, v_\alpha)$ indicates that variable $f_\alpha$ was measured with value $v_\alpha$ at time $t_\alpha$. To enhance comparability, all values are standardized on a per-variable basis.

To prepare the data for the transformer, we bin the time series into hourly buckets, retaining only the first observed value per hour and variable. 
Given the high sparsity of the data, we apply zero-mean imputation and simultaneously generate a masking matrix indicating which values were missing.

Next, we apply an encoder embedding function to transform the samples into embeddings, which are subsequently used for data augmentation because they provide a consistent matrix size without missing or imputed data. This approach exclusively focuses on the right side of Figure~\ref{fig:neural-architecture}, where the embeddings of the original data are fixed, as are the weights of the dense encoder embedding function. Generating new data begins at the embedding layer, making this strategy particularly valuable for enhancing model robustness and improving generalization without compromising data integrity. 

The embeddings are processed by an encoder, followed by an autoregressive Transformer decoder for forecasting.
For the decoder, we employ an Iterative Multi-Step (IMS) approach, where an output vector $\hat{y} \in \mathbb{R}^{|F|}$ is generated using the history of previous predictions. 

For long-term TSF, we concatenate the predictions over all time steps $t = 1, \ldots, T$. As demonstrated in \cite{StaniekETAL:24mlhc}, the autoregressive decoder performs better when applied with student forcing, where previous predictions are treated as ground truth during training.

\paragraph{Performance Evaluation Metrics}

A data point $x = (e,y,m)$ representing a patient's ICU stay consists of an embedding matrix $e \in \mathbb{R}^{24\times n}$ representing the first 24 hours, a ground truth time series $y \in \mathbb{R}^{T\times |F|}$ for the subsequent $T$ hours, and a mask $m \in \{0,1\}^{T\times |F|}$  indicating the presence of measurements in $y$ with the count $|m|$. $T$ indicates the number of hours predicted. The dimension $n$ of the embedding is model-dependent. 

To evaluate the performance of TSF models, we define a masked mean squared error by
	\begin{equation}\label{eq:MSE}
		\text{mMSE}(x,\theta) := \frac1{|m|} \| (f_\theta (e)-y) \odot m\|_2^2
	\end{equation}
    for a single sample. The mMSE is masked to ensure that only observed values contribute to the loss calculation, effectively ignoring missing data points. It is then normalized by the cardinality of the mask, enhancing the comparability between samples with varying amounts of observed data. For a dataset $X$, we generalize $\text{mMSE}$ to
	\begin{equation}\label{eq:MSE_set}
		\text{MSE}(X,\theta) := \frac1{|X|} \sum_{x\in X} \text{mMSE}(x,\theta).
	\end{equation}

\section{Membership Inference Attack (MIA) and Privacy Loss Evaluation}

The loss-based MIA described in \cite{YeomETAL:18} can be formalized as follows: 

\begin{definition}[Loss-based MIA]\label{def:MIA}
A loss-based MIA is a privacy attack on a machine learning model $f$ where an adversary $\mathcal{A}$ attributes a positive membership to a datapoint $x$ if the model loss on the datapoint $\ell(f(x),y)$ is lower than a threshold $\tau$ for a model prediction $f(x)$ and a label $y$:
\begin{align}
    \mathcal{A}_{\text{loss}}(x,y) = \mathbb{I}_{\ell(f(x),y) < \tau}
\end{align}
\end{definition}

In order to evaluate the attacker's success, we follow \cite{CarliniETAL:22} who replace dataset-wide evaluation metrics for success rates of privacy attacks by an evaluation that measures if a MIA can reliably violate the privacy, even if it affects just a few users. Conversely, an attack that unreliably achieves a high aggregate success rate should not be considered successful. This is achieved by an evaluation of the attacker's True Positive Rate (TPR) at low False Positive Rate (FPR). 

\paragraph{TPR/FPR ratio} We assume that the attacker has access to both mMSE (Equation~\ref{eq:MSE}), for example, via an API that provides access to the model as scoring device, and to MSE (Equation~\ref{eq:MSE_set}), for example, via published results on the model's average MSE on particular datasets. To defend against a loss-based attack with $\tau$ set to the average training loss, we aim for a low TPR and a high FPR for an attacker $\mathcal{A}$. TPR and FPR are defined as follows: 
\begin{align}
    \text{TPR}(X_\text{train},\tau, \theta) &:= \frac1{|X_\text{train}|} \sum_{x \in X_\text{train}}\mathbb{I}_{\text{mMSE}(x, \theta)<\tau}\\
    \text{FPR}(X_\text{test},\tau, \theta) &:= \frac1{|X_\text{test}|} \sum_{x \in X_\text{test}}\mathbb{I}_{\text{mMSE}(x, \theta)<\tau}
\end{align}

In other words, members should ideally not be recognized as members (minimizing TPR), while non-members should be classified as members (maximizing FPR) until the ratio reaches 1 and non-members become indistinguishable from members. This TPR/FPR ratio is advantageous as it remains valid even for unbalanced test sets. Following \cite{YeomETAL:18}, we assume the attacker is aware of the average training loss, which we will use as threshold $\tau$. This yields the following privacy metric:

	\begin{equation}\label{eq:priv}
		\text{Priv}(X_\text{train},X_\text{test},\tau, \theta) := \frac{\text{TPR}(X_\text{train},\tau, \theta)}{\text{FPR}(X_\text{test},\tau, \theta)}
	\end{equation}
This metric represents the attacker's advantage in correctly identifying members compared to incorrectly identifying non-members. A lower Priv value indicates less advantage for the attacker.

Additionally, we define privacy loss (PL) as the positive membership prediction of a single data point:
	\begin{equation}\label{eq:PL}
		\text{PL}(x, \tau, \theta) := \mathbb{I}_{\text{mMSE}(x, \theta)<\tau}
	\end{equation}

In our implementation, the threshold $\tau$ is set to the average training loss, computed after each training run using the current model parameters $\theta$. This ensures that $\tau$ adapts to each model instance and reflects the actual distribution of losses encountered during training.

\paragraph{ROC curve}
By varying the threshold $\tau$ and calculating the corresponding TPR and FPR values, we generate the Receiver Operating Characteristic (ROC) curve. The area under the ROC curve (AUROC) provides a quantitative measure of the model's susceptibility to membership inference. An AUROC value of 0.5 indicates that the training and held-out sets are indistinguishable, implying no vulnerability to the attack. Conversely, an AUROC value approaching 1 suggests that the sets are highly separable, indicating a significant privacy risk.

\section{Data Augmentation}
\label{sec:augment}

The central idea of data augmentation is to improve a model's generalization performance by re-training on a dataset augmented with synthesized examples that cover aspects of the data distribution that are not found in the training sample. In our work, we focus on data augmentation techniques that synthesize data in the space of neural embeddings. We compare mix-based data augmentation \citep{ZhangETAL:18mixup,CaoETAL:22} with zeroth-order optimization in embedding space \citep{NesterovSpokoiny:15,ChenETAL:17}. We use data augmentation not only to reduce MSE on the held-out set but also to lower the TPR/FPR ratio of the attacker. In our experiments, new data is generated within each epoch and the model is trained on a mix of original training data and synthetic data, while the embedding layer is kept fixed.

\paragraph{ZOO}

Zeroth-order optimization methods are of interest for machine learning problems where only the zeroth-order oracle, i.e., the value of the objective function but no explicit gradient, is available. Our application of ZOO is inspired by a black-box attack on deep neural networks where adversarial images that lead to misclassification are found by approximating the gradient through a comparison of function values at random perturbations of input images \citep{ChenETAL:17} or language \citep{BergerETAL:21}. In addition to being simple and scalable, ZOO methods can be adjusted to optimize non-differentiable functions and are still provably convergent  \citep{Fu:06,NesterovSpokoiny:15}. 

The central idea of ZOO is to perform optimization in the space of neural embeddings by iteratively updating an embedding for $S$ steps by evaluating an objective function $g$ over randomly perturbed points. This procedure includes the following iterative steps:

\begin{enumerate}
    \item Generate $k$ random perturbations $u_i$ drawn from a normal distribution $\mathcal{N}(0, \mathbb{I})$ and normalized to $\|u_i\| = 1$, $i= 1,\dots,k$.
    \item Given a data point $x = (e,y,m)$, consisting of an embedding matrix $e \in \mathbb{R}^{24\times n}$, a ground truth time series $y \in \mathbb{R}^{T\times |F|}$, and a mask $m \in \{0,1\}^{T\times |F|}$, compute perturbed samples $x_i$:
    \[
    x_i^\pm = ({e}_s \pm \mu u_i, m, y)
    \]
    \item Update the embedding matrix using:
    \[
    {e}_{s+1} = {e}_s - \lambda \frac{1}{k} \sum_{i=1}^{k} \frac{g(x_i^+, X, \theta) - g(x_i^-, X, \theta)}{2 \mu} u_i
    \]
\end{enumerate}

\begin{itemize}
    \item $\lambda \in \mathbb{R}$: Learning rate controlling the step size of updates.
    \item $\mu \in \mathbb{R}$: Perturbation width controlling the scale of sampling around ${e}_s$.
\end{itemize}

We use a parameterized objective function $g_\alpha$ that is designed to improve both utility and privacy of medical data. Utility is optimized by searching for embedding matrices that represent examples that are sufficiently different from the training examples, in order to allow a decrease in held-out loss after re-training. This is achieved by using a negative mMSE term (Equation~\ref{eq:MSE}) in the loss objective. Privacy protection is optimized by searching for embedding matrices that are similar to the training examples in order to distract the attacker. This is achieved by using a negative PL term (Equation~\ref{eq:PL}) in the loss objective. The interpolation weight $\alpha \in [0,1]$ balances these two competing goals: when $\alpha = 1$, we purely optimize for diversity (high MSE on synthetic samples encourages exploration); when $\alpha = 0$, we purely optimize for privacy (low PL makes synthetic samples indistinguishable from training data). In practice, intermediate values of $\alpha$ achieve the best utility-privacy tradeoff:

\begin{equation}\label{eq:objective_function_g}
    g_\alpha(x, X, \theta) = -\Big(\alpha \,\text{mMSE}(x, \theta) + (1 - \alpha) \,\text{PL}(x, X, \theta)\Big)
\end{equation}

\paragraph{ZOO-PCA}

In traditional ZOO, the perturbations are generated randomly from a standard normal distribution, meaning that each feature in the embedding matrix is perturbed independently. While this can help explore the parameter space, it may not always result in meaningful or efficient perturbations, especially when certain features are more important than others for the task at hand.

In ZOO-PCA, by performing Principal Component Analysis (PCA) first, we are essentially identifying the most significant directions in the data that explain its variance. These principal components represent the major patterns or features of variation in the data, rather than treating all features equally. By keeping only the components that explain a significant portion of the variance (based on the cumulative explained variance ratio threshold), ZOO-PCA guides perturbations toward the most relevant variance components of the data, rather than random or less important components.

\paragraph{MixUp}

The MixUp technique of \cite{ZhangETAL:18mixup} generates a new data point by interpolating between two randomly selected data points \( x_1 = (e_1, y_1, m_1) \) and \(x_2 = (e_2, y_2, m_2) \).
With a $\lambda \sim \text{Beta}(\beta,\beta)$, we receive:

\begin{equation}
    x_\text{MixUp1,2} = 
    \begin{cases} 
        \big(\lambda e_1 + (1 - \lambda) e_2, y_1, m_1\big), & \text{if } \lambda > 0.5 \\
        \big(\lambda e_1 + (1 - \lambda) e_2, y_2, m_2\big), & \text{if } \lambda \leq 0.5
    \end{cases}
\end{equation}

The new data point has a new embedding matrix, and the mask and target of the dominant data point. Due to the high sparsity of medical time series this is a good way to have a working perturbation instead of mixing most of the time \textit{zero values} with observations.
With a small $\beta$ ($\beta \ll 1$), the new data points resemble one example more closely, while with a large $\beta$ ($\beta \gg 1$), the examples are pushed toward a mixture of both points.

\begin{figure}[t!]
    \centering
    \includegraphics[width=0.92\columnwidth]{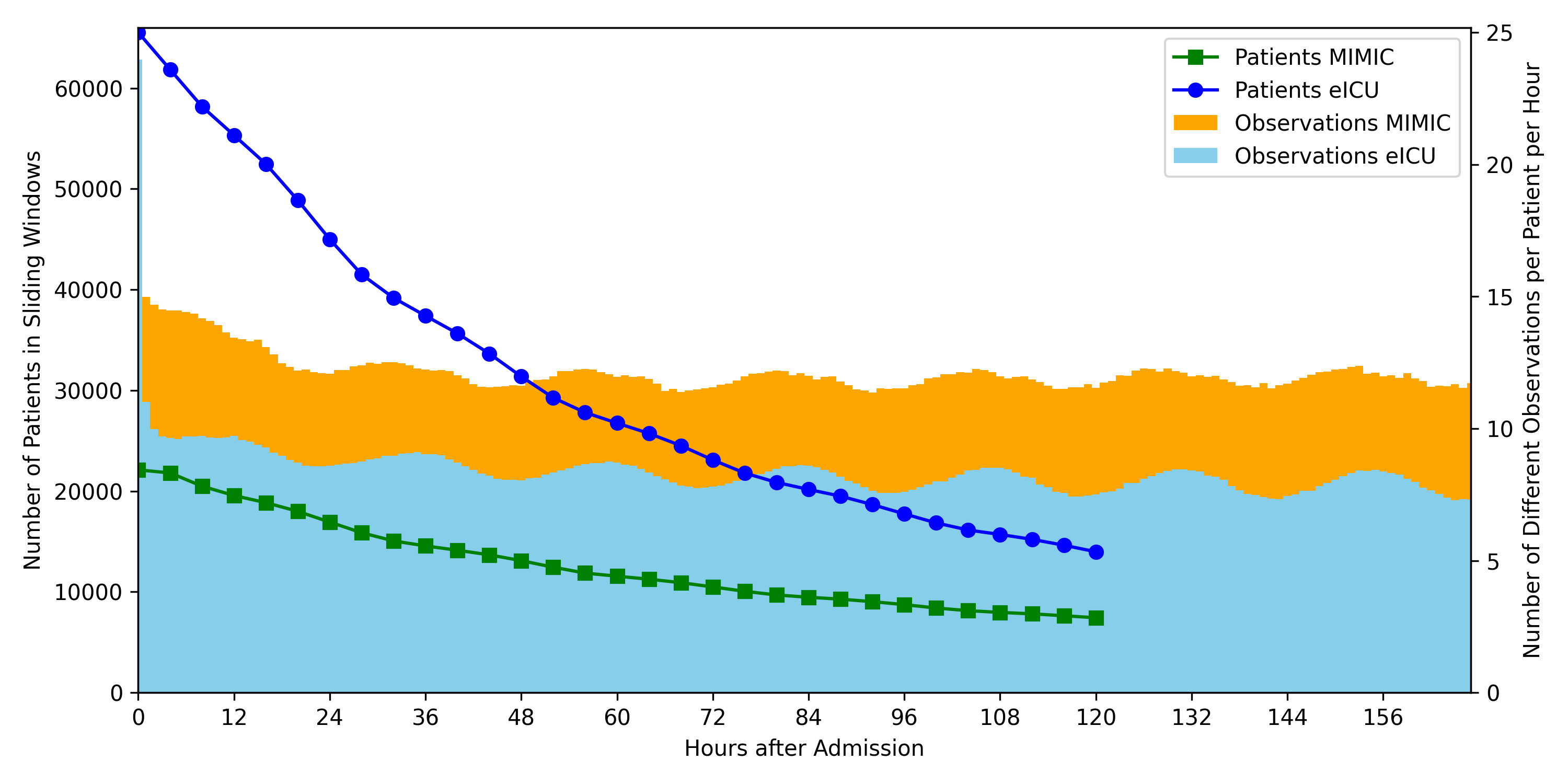}
    \caption{
    The bar plots (right axis) show the average number of different clinical variables recorded per patient per hour after admission (multiple measurements of the same variable within one hour are deleted during binning). These counts remain relatively stable over time for both MIMIC (orange) and eICU (blue). In contrast, the line plots (left axis) display the number of patients contributing to each 4-hour sliding window, which declines over time -- more sharply in eICU -- reflecting the decreasing number of long-staying patients.
    }
    \label{fig:sliding-window}
\end{figure}

\section{Experimental Setup}

In our experiments, we use the Medical Information Mart for Intensive Care III (MIMIC-III) data \citep{JohnsonETAL:16} and the larger eICU dataset \citep{PollardETAL:18}. A detailed description of the data is given in Appendix \ref{app:data}.

\begin{table*}[t!] \centering 
\caption{Evaluation results for baseline and data augmentation methods for TSF on MIMIC-III and eICU. All results are reported on unseen test data. Generalization performance of models is reported as MSE on test data MSE($X_\text{test}$). Effectiveness of a loss-based MIA is reported by TPR/FPR ratio using the average training loss as threshold. An approximate randomization test \citep{Noreen:89} shows that all result differences are significant (with $p<0.05$) except for the MSE between baseline and ZOO-PCA for the eICU dataset. Best results for data augmentation methods are shown in bold face. FPR@$\tau$ values show that ZOO-based methods and DP-SGD increase false positive rates, effectively distracting the attacker.} \label{tab:results} 
\begin{tabular}{lccc|ccc} \toprule 
& \multicolumn{3}{c}{MIMIC-III} & \multicolumn{3}{c}{eICU} \\  
\cmidrule(lr){2-4} \cmidrule(lr){5-7} 
Method & MSE($X_\text{test}$) & $\frac{TPR}{FPR}@\tau$ & ${FPR}@\tau$ &MSE($X_\text{test}$) & $\frac{TPR}{FPR}@\tau$ & ${FPR}@\tau$ \\ 
\midrule 
Baseline & 0.5053 & 3.5482 & 0.1870& 0.5604 & 1.3206 &0.5137\\ 
DP-SGD &0.7446	&0.9887	& 0.6560&0.8228&0.9969 & 0.6676\\ 
\midrule
ZOO & 0.4945 & 1.9925 & 0.4148 &0.5576 & 1.1716 & 0.6877\\ 
ZOO-PCA & 0.5000 & \textbf{1.4271} & 0.6421 & 0.5607 & \textbf{1.1299} & 0.7712 \\ 
MixUp & \textbf{0.4918} & 3.2264 & 0.1936& \textbf{0.5519} & 1.2927 & 0.5000 \\ 
\bottomrule 
\end{tabular}
\end{table*}

In all of our experiments, we go beyond the standard forecasting setup -- predicting hours 24 to 48, based on input from hours 0 to 24 -- by employing a sliding window technique that covers a wider range of temporal patterns. Specifically, for each patient, we shift both the observation and prediction windows in four-hour intervals throughout the ICU stay, enabling forecasts such as hours 40 to 64 based on inputs from hours 16 to 40. This strategy allows us to capture temporal dynamics more comprehensively and is illustrated in Figure~\ref{fig:sliding-window}. As a result, we do not restrict our analysis to the initial data points but include all windowed segments up to the first four days after admission, while the set splits stay on patient level. This significantly expands our datasets: from 21,573 to 342,213 samples in MIMIC-III (a 15.9-fold increase) and from 65,053 to 868,487 in eICU (a 13.4-fold increase). 

For each data augmentation run, we generate 32,000 new synthetic examples. After the generation process, the model is re-trained on a balanced dataset composed of 50\% original and 50\% synthetic data. Due to memory constraints, the number of synthetic examples is limited such that it never exceeds 50\% of the size of the training set. If this limit is reached, the oldest synthetic samples are discarded to make room for newly generated ones.

The updated model parameters are denoted by $\theta$, and the augmented dataset $X_\text{aug}$ comprises both the original training data $X_\text{train}$ and the generated synthetic data. The mean squared error (MSE; see Equation~\ref{eq:MSE_set}) is evaluated on a held-out set using the current model parameters. The privacy metric (Equation~\ref{eq:priv}) treats the training set as the member set, the held-out set as the non-member set, and uses the augmented set as the reference for computing the average loss $\tau$.

For all training runs, the model is only updated if the following conditions are satisfied:

\begin{align*}
    & \text{Priv}(X_\text{train}, X_\text{heldout}, \tau, \theta) \leq  (1+\varepsilon_\text{priv}) \, \text{Priv}_\text{best},  \\
    & \text{MSE}(X_\text{heldout}, \theta) \leq (1+\varepsilon_\text{MSE}) \, \text{MSE}_\text{best},  \\
    & \text{Priv}(X_\text{train}, X_\text{heldout}, \tau, \theta) + \beta \, \text{MSE}(X_\text{heldout}, \theta) \\
    & \leq  \text{Priv}_\text{best} + \beta \, \text{MSE}_\text{best}.
\end{align*}

These criteria ensure that the model is not updated unless it preserves a balance between utility and privacy: neither the privacy risk nor the held-out MSE may increase beyond a small threshold, and the combined metric must improve. Thus, the optimization does not aim for absolute minima in either privacy or utility alone, but instead for an improved tradeoff between both. In our experiments, we set $\varepsilon_\text{priv} = \varepsilon_\text{MSE} = 0.5\%$ and $\beta = 3$.
Complete meta-parameter settings are given in Appendix~\ref{app:metaparameter}.

\begin{figure}[t!]
    \centering
    \includegraphics[width=0.84\columnwidth]{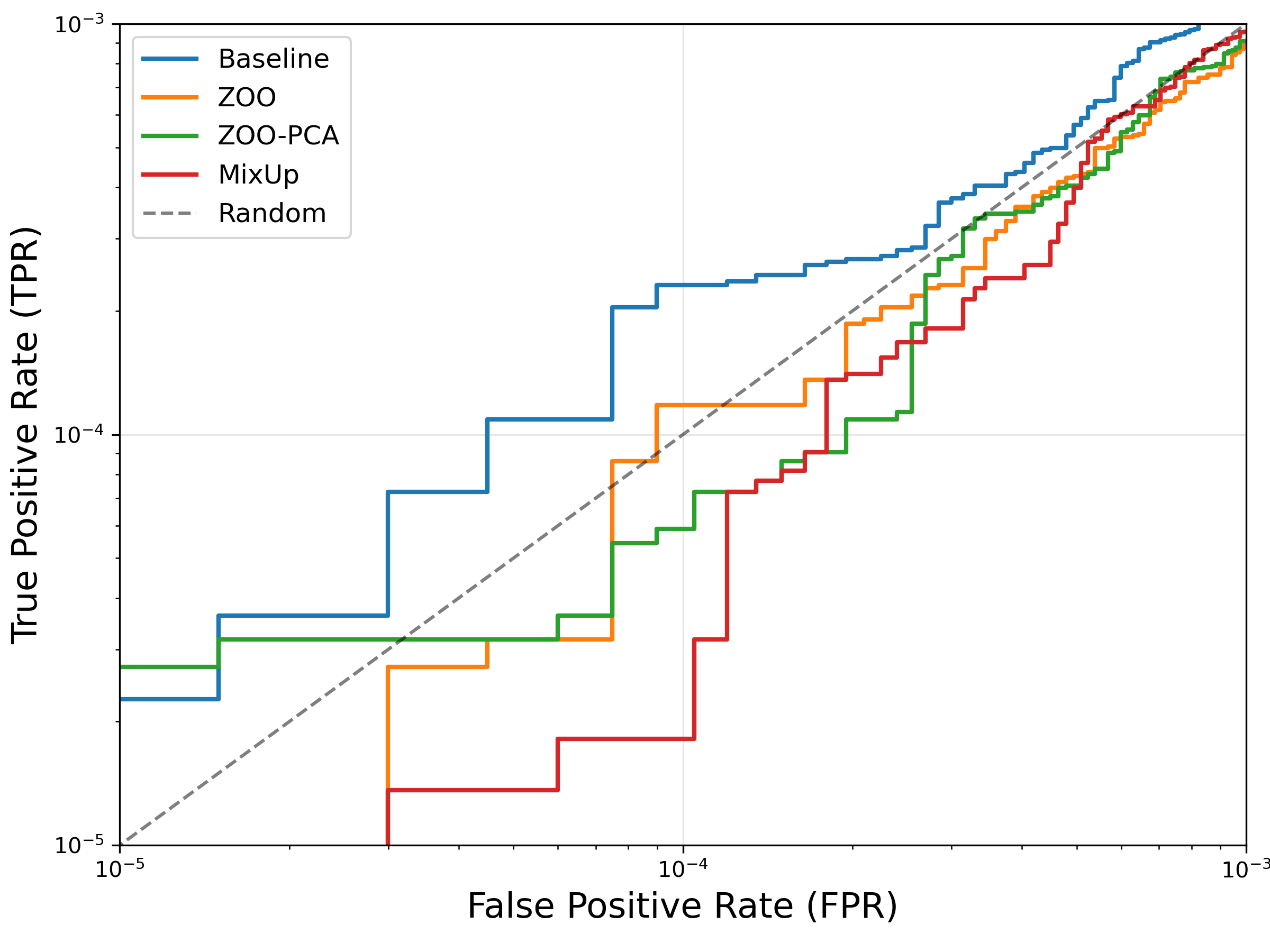}
    \includegraphics[width=0.84\columnwidth]{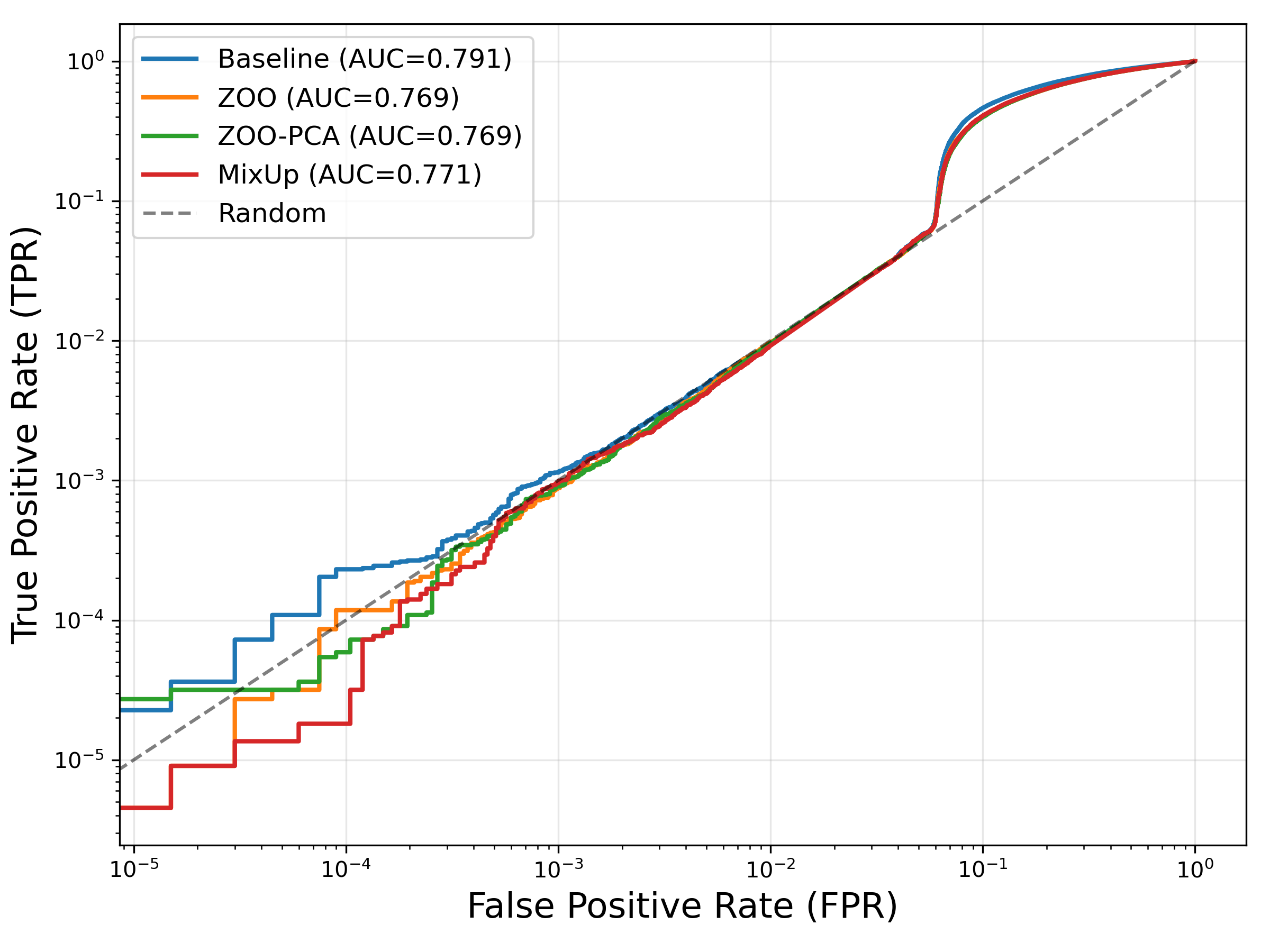}
    \caption{MIMIC-III: ROC curves (log-log scaling) for varying thresholds of loss-based MIAs on models trained with and without data augmentation. Upper plot magnifies the area for FPR $<$ 0.1\%. The DP-SGD curve (not shown) is nearly indistinguishable from the diagonal, representing random guessing.}
    \label{fig:auroc-mimic}
\end{figure}

\section{Experimental Results}

In our experiments, all results are reported on unseen test data. The evaluation metric used is MSE (Equation~\ref{eq:MSE_set}) on test data to measure the generalization performance. The effectiveness of the loss-based MIA described in Definition \ref{def:MIA} is measured by the TPR/FPR ratio with $\tau$ set to the average training loss, and negative examples taken from the test set. Results are reported in Table~\ref{tab:results} for the best tradeoff between privacy protection and utility on validation data for each method, respectively. 

\begin{figure}[t!]
    \centering
    \includegraphics[width=0.84\columnwidth]{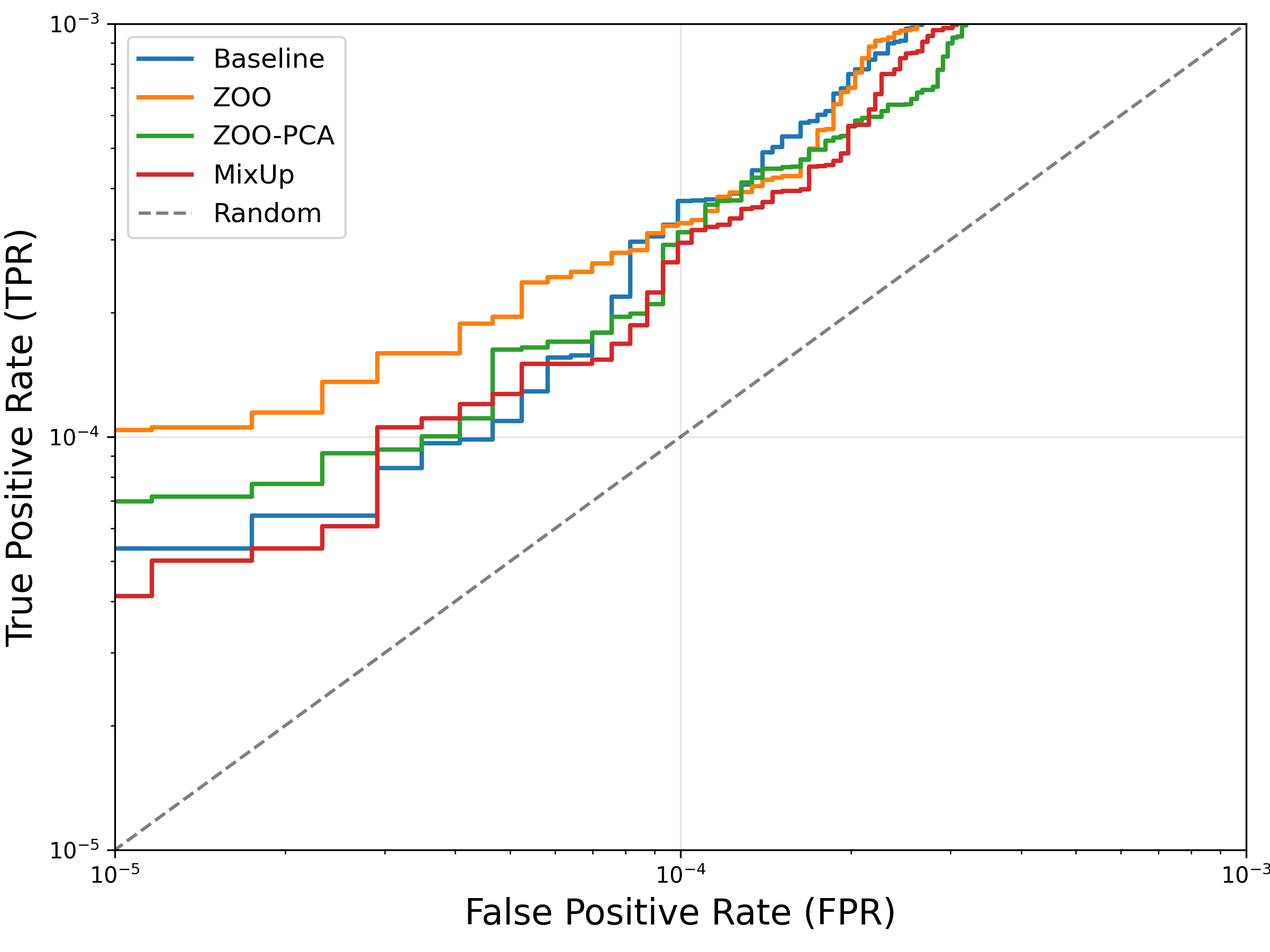}
    \includegraphics[width=0.84\columnwidth]{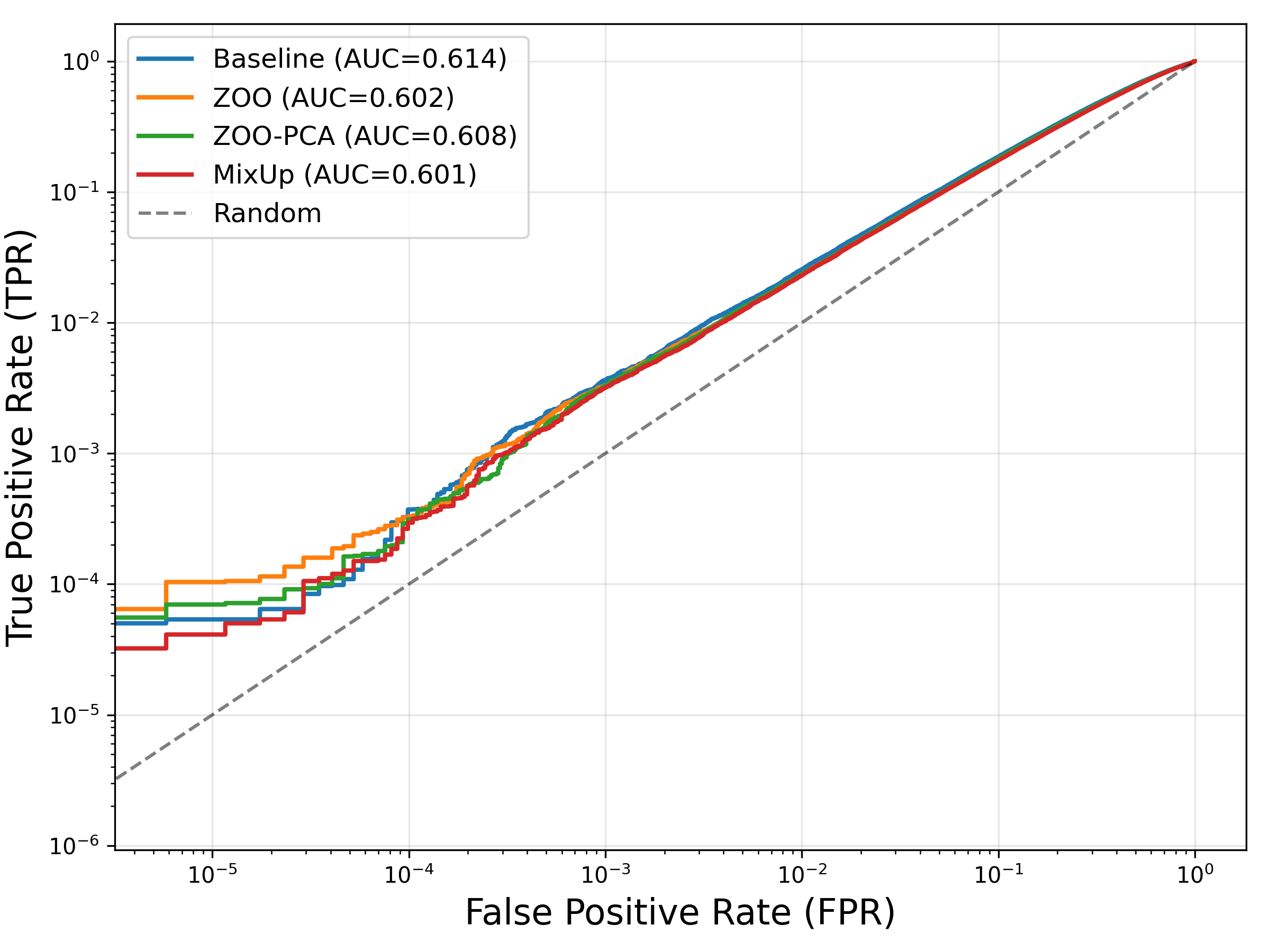}
    \caption{eICU: ROC curves (log-log scaling) for varying thresholds of loss-based MIAs on models trained with and without data augmentation. Upper plot magnifies the area for FPR $<$ 0.1\%. The DP-SGD curve (not shown) is nearly indistinguishable from the diagonal, representing random guessing.}
    \label{fig:auroc-eicu}
\end{figure}

We present two complementary evaluations of MIA's effectiveness: Table~\ref{tab:results} follows \citeauthor{YeomETAL:18}'s \citeyearpar{YeomETAL:18} fixed-threshold scenario where $\tau$ is set to the average training loss (the only information realistically available to an attacker), while Figures~\ref{fig:auroc-mimic} and \ref{fig:auroc-eicu} show \citeauthor{CarliniETAL:22}'s \citeyearpar{CarliniETAL:22} recommended evaluation using ROC curves with varying thresholds, particularly focusing on the low FPR regime where privacy breaches are most critical.

As baseline we use a model that is trained to convergence on the training set without the use of synthetic data. 

Furthermore, we compare to DP-SGD \citep{AbadiETAL:16}, a standard method to foster differential privacy by clipping and adding Gaussian noise to the gradient updates. Following \cite{AbadiETAL:16}, the privacy accounting (moments accountant) provides meaningful $\varepsilon$-DP bounds primarily for $\sigma > 1$. We implemented DP-SGD with noise multipliers ranging from $\sigma = 1.1$ to $\sigma = 10$, gradient clipping norms $C \in \{1.1, 1.5, 2.0\}$, and increased the learning rate by 100 against the baseline to compensate for the added noise.  While DP-SGD achieves strong privacy protection with TPR/FPR ratios approaching 1.0 at any noise levels, this comes at the cost of severe utility degradation. 
Table~\ref{apd:dp_sigma} shows that even at the lowest noise level ($\sigma=1.1$), DP-SGD achieves strong privacy protection (TPR/FPR $\approx 1$) but at significant cost to utility, with MSE degrading by 47\% on MIMIC-III (0.5053) and 47\% on eICU (0.5604) compared to baseline. This demonstrates that DP-SGD's noise injection mechanism is poorly suited to high-dimensional, sparse EHR forecasting tasks.
 While DP-SGD can provide theoretical guarantees during the augmentation phase, it is important to note that our overall system cannot provide end-to-end $\varepsilon$-DP guarantees. This is because our pretrained embedding layer was trained without DP. By the composition theorem \citep{Dwork:06}, $\varepsilon_\text{total} = \varepsilon_\text{pretrain} + \varepsilon_\text{augmentation}$, and since $\varepsilon_\text{pretrain}$ is unbounded, the partial privacy bound has limited practical value in our specific setup.
 For the results in Table~\ref{tab:results}, we used $\sigma = 1.1$ and $C = 2$ with a learning rate of $0.005$.

\begin{table}[t!]
\centering
\caption{Evaluation of DP-SGD with varying noise multipliers ($\sigma$) and fixed clipping norm $C=2$ on MIMIC-III and eICU datasets.}
\label{apd:dp_sigma}
\resizebox{0.495\textwidth}{!}{
\begin{tabular}{c|cc|cc}
\toprule
& \multicolumn{2}{c|}{MIMIC-III} & \multicolumn{2}{c}{eICU} \\
\cmidrule(lr){2-3} \cmidrule(lr){4-5}
$\sigma$ & MSE($X_\text{test}$) & $\frac{TPR}{FPR}@\tau$ & MSE($X_\text{test}$) & $\frac{TPR}{FPR}@\tau$ \\
\midrule
1.1 & 0.7446 & 0.9887 & 0.8228 & 0.9969 \\
1.5 & 0.7655 & 0.9913 & 0.9324 & 0.9976 \\
2.0 & 0.8403 & 0.9844 & 0.9938 & 0.9945 \\
3.0 & 0.7952 & 0.9838 & 1.0610 & 0.9929 \\
4.0 & 0.7788 & 0.9816 & 1.0302 & 0.9952 \\
5.0 & 0.7900 & 0.9823 & 0.9931 & 0.9942 \\
6.0 & 0.8044 & 0.9795 & 0.9835 & 0.9966 \\
7.0 & 0.8150 & 0.9861 & 0.9774 & 0.9963 \\
8.0 & 0.8320 & 0.9814 & 0.9762 & 0.9951 \\
\bottomrule
\end{tabular}
}
\end{table}

The data augmentation methods are ZOO, ZOO-PCA, and MixUp as described in Section \ref{sec:augment}. The MSE($X_\text{test}$) results in Table~\ref{tab:results} show that all data augmentation techniques improve generalization performance on test data compared to the baseline, while best results are achieved for MixUp due to its ability to explore distant data regions while ZOO stays close to training points. This shows MixUp to be a strong technique to improve generalization by data augmentation. The TPR/FPR ratio results in Table~\ref{tab:results} show that all data augmentation methods achieve a decrease in the effectiveness of the loss-based MIA, with best results obtained by the ZOO-PCA technique.  In our experiments, we keep as many principal components as necessary to explain 70\% of the variance. This threshold was found to be optimal by experiments on the held-out set testing 50\% to 90\% (see Appendix~\ref{app:results_zoopca}). This shows that the setting of the ZOO algorithm that prefers synthesis of complex data in order to increase the FPR of the attacker, together with a focus on principal components in embedding space, yields the best tradeoff between generalization and privacy protection.

Figures~\ref{fig:auroc-mimic} and \ref{fig:auroc-eicu} show the ROC curves for the loss-based MIA for varying thresholds $\tau$. We see that especially at low FPR values, the TPR/FPR ratio is high for the baseline model, and substantially reduced for the models trained with data augmentation. The ZOO data augmentation method forces the ROC curve of the attacker close to the diagonal, which corresponds to an attacker based on random choice.

 For a visual analysis of the privacy-utility tradeoff across different augmentation parameters, see Appendix~\ref{app:visu}.

\section{Discussion}

Our study demonstrates the effectiveness of data augmentation techniques, particularly zeroth-order optimization, and its PCA-restricted variant, in optimizing both utility and privacy in medical time series forecasting. In fact we found that all data augmentation methods improve privacy against MIA at similar or smaller MSE values on test data, albeit not by huge amounts. In contrast, noise injection on gradients as done in DP-SGD requires noise levels that destroy the model's ability to learn useful patterns. The results presented in Table~\ref{tab:results}  highlight a significant reduction in the effectiveness of loss-based membership inference attacks through the strategic generation of synthetic data. 
The ROC curves in Figures~\ref{fig:auroc-mimic} and \ref{fig:auroc-eicu} emphasize the privacy benefits of our approach. At low FPR values, where the risk of privacy breaches is most critical, models trained with ZOO-based augmentation significantly outperform the baseline model. The near-diagonal ROC curve achieved by ZOO indicates that the attacker's performance is close to random guessing, effectively mitigating the risk of MIA.

The ZOO-PCA method, which perturbs embeddings along the directions of principal components, proved particularly effective in reducing the TPR/FPR ratio. We hypothesize that ZOO-PCA directs the augmentation process along the most significant data variations, whereas random perturbations could lead to embeddings that are outside the convex hull of the input embedding space. Since ZOO-PCA effectively performs sparse zeroth-order optimization, it enjoys favorable convergence rates \citep{BalasubramanianETAL:22}. Furthermore, directing the augmentation process along the most significant data variations can better confuse the attacker while maintaining or even improving model utility. In contrast, MixUp, a randomization-based approach, demonstrated strong generalization capabilities, achieving the lowest MSE on the held-out set. This highlights the complementary strengths of gradient-free optimization and randomization techniques in data augmentation.

\section{Conclusion}

Our study demonstrates that embedding-space data augmentation can effectively mitigate membership inference attacks while preserving predictive performance in clinical time series forecasting. ZOO-PCA achieves the best privacy-utility tradeoff, while MixUp excels at generalization, highlighting complementary strengths of different augmentation strategies.
In future work we intend to explore hybrid approaches and investigate applicability to other deep learning architectures and privacy attack scenarios.

\acks{
We acknowledge support by the state of Baden-Württemberg through bwHPC
and the German Research Foundation (DFG) through grant INST 35/1597-1 FUGG.

The first author was supported by the Helmholtz Association under the joint research school HIDSS4Health — Helmholtz Information and Data Science School for Health.}

\bibliography{ref}

\clearpage
\appendix

\section{Data}
\label{app:data}

\paragraph{MIMIC-III:} MIMIC-III was collected from the Beth Israel Deaconess Medical Center between 2001 and 2012 and contain over 40k patients. After filtering for patients with an ICU stay of at least 24 hours with reported gender and age of at least 18 years, our dataset contains 44,858 ICU stays with 56 million data points. We split the data into partitions for training (28,791), held-out (7,144), and test (8,878). For our study, we used 131 different clinical variables. The full list of extracted MIMIC-III features is given in Appendix~\ref{app:mimic-features}. 

Converting the data to a dense one hour representation yields 89.08\% missing data, changing per variable from under 15\% (HR, RR, SBP, DBP, MBP, and O2 Saturation) to more than 90\% for 101 variables, and exceeding 99\% for 42 variables. On the other side we are losing 17.73\% of the data points through the densification procedure where multiple measurements occur within the same hour and for the same variable. 
The high sparsity of the MIMIC-III data presents a significant challenge for time series forecasting, requiring careful handling of missing values and the development of robust models.

\paragraph{eICU:} The eICU data was collected from over 200 US hospitals and comprise over 200,000 ICU stays. After filtering for patients with an ICU stay of at least 48 hours, reported gender and aged 18 years or older, 
we arrived at 77,704 ICU stays with 415 million data points. This set was partitioned in subsets for training (49,730), development (12,433), and testing (15,541).  As shown in Appendix \ref{app:eICU-features}, we extracted 98 clinical variables for our experiments. The measurements in the eICU data set are denser than in MIMIC-III since the number of observations per patient per hour is three times higher than for MIMIC-III and decreases at a slower rate with length of stay, also there are a lot of bed values.
After our binning process, this reduces to one sixth of the measurements (16.86\%) leading to still  89.85\% missing data. The same six variables (HR, RR, SBP, DBP, MPB, and O2 Saturation) as before are quite complete.

\newpage

\section{Meta-parameters}
\label{app:metaparameter}

The following meta-parameter settings were used in our experiments:

\begin{table}[h!]
\caption{Hyperparameter configurations}
\centering
ZOO and ZOO-PCA parameters

\begin{tabular}{rrrr}
\toprule
$\lambda$ & $\mu$ & k & \texttt{steps} \\
\midrule
3000 & 300 & 3 & 10 \\
\bottomrule
\end{tabular} \vspace{1em}

Variants of ZOO/ZOO-PCA ($\alpha$) and MixUp ($\beta$)

\begin{tabular}{cc}
\toprule
$\alpha$ & $\beta$ \\
\midrule
\{0, 0.25, 0.5, 0.75, 1\} & \{0.2, 1, 5\} \\
\bottomrule
\end{tabular} \vspace{1em}

Acceptance Criteria

\begin{tabular}{ccc}
\toprule
$\varepsilon_\text{priv}$ & $\varepsilon_\text{MSE}$ & $\beta$ \\
\midrule
 0.5\% & 0.5\% & 3\\
\bottomrule
\end{tabular}\vspace{1em}

DP-SGD parameter

\begin{tabular}{cc}
\toprule
Noise Multiplier & Clipping Norm \\
\midrule
\{1.1, 1.5, 2\} & \{1.1, 1.5, 2\} \\
\{2\} & \{3,4,5,6,7,8,16,50\} \\
\bottomrule
\end{tabular}

\end{table}

\section{Evaluation of ZOO-PCA variance explanation}
\label{app:results_zoopca} 
\begin{table}[!htbp] \centering \caption{Evaluation results for baseline and PCA ZOO data augmentations. The most fitting variance explanation parameter was selected via grid search on validation data from $\{ 50\%,70\%, 90\%\}$. Compare to Table~\ref{tab:results}.
} \label{tab:results_zoopca} 
\resizebox{0.99\columnwidth}{!}{
\begin{tabular}{lcc|cc} \toprule 
& \multicolumn{2}{c}{MIMIC-III} & \multicolumn{2}{c}{eICU} \\  
\cmidrule(lr){2-3} \cmidrule(lr){4-5} 
PCA ratio & MSE($X_\text{test}$) & $\frac{TPR}{FPR}@\tau$ & MSE($X_\text{test}$) & $\frac{TPR}{FPR}@\tau$ \\ 
\midrule 
Baseline & 0.5053 & 3.5479 & 0.5604 & 1.3206 \\ \midrule 
50\% & 0.5037 & 1.713 & 0.5602& 1.1526\\
70\% & 0.5000 & \textbf{1.4271} & 0.5607 & \textbf{1.1299} \\ 
90 \% & \textbf{0.4994} & 1.455 & \textbf{0.5597}& 1.1612\\
\bottomrule 
\end{tabular}
}
\end{table}

\onecolumn 
\clearpage

\section{MIMIC-III features}
\label{app:mimic-features}

\begin{table}[!h]
    \centering
    \caption{For MIMIC-III, 131 dynamic variables were extracted.}
    \label{tab:features_mimic}
\begin{tabular}{llll}

\toprule
ALP                  & Epinephrine          & LDH                       & Packed RBC           \\
ALT                  & Famotidine            & Lactate                   & Pantoprazole         \\
AST                  & Fentanyl              & Lactated Ringers          & Phosphate            \\
Albumin              & FiO2                 & Levofloxacin              & Piggyback            \\
Albumin 25\%         & Fiber                 & Lorazepam                 & Piperacillin         \\
Albumin 5\%          & Free Water            & Lymphocytes               & Platelet Count      \\
Amiodarone           & Fresh Frozen Plasma   & Lymphocytes (Absolute)    & Potassium            \\
Anion Gap            & Furosemide            & MBP                       & Pre-admission Intake \\
BUN                  & GCS\_eye             & MCH                       & Pre-admission Output \\
Base Excess          & GCS\_motor           & MCHC                      & Propofol             \\
Basophils            & GCS\_verbal          & MCV                       & RBC                  \\
Bicarbonate          & GT Flush              & Magnesium                 & RDW                  \\
Bilirubin (Direct)   & Gastric               & Magnesium Sulfate (Bolus) & RR                   \\
Bilirubin (Indirect) & Gastric Meds          & Magnesium Sulphate        & Residual             \\
Bilirubin (Total)   & Glucose (Blood)       & Mechanically ventilated   & SBP                 \\
CRR                  & Glucose (Serum)       & Metoprolol                & SG Urine             \\
Calcium Free         & Glucose (Whole Blood) & Midazolam                 & Sodium               \\
Calcium Gluconate    & HR                    & Milrinone                 & Solution             \\
Calcium Total        & Half Normal Saline    & Monocytes                 & Sterile Water        \\
Cefazolin            & Hct                   & Morphine Sulfate          & Stool                \\
Chest Tube           & Heparin               & Neosynephrine             & TPN                  \\
Chloride             & Hgb                   & Neutrophils               & Temperature          \\
Colloid              & Hydralazine           & Nitroglycerine            & Total CO2            \\
Creatinine Blood    & Hydromorphone         & Nitroprusside             & Ultrafiltrate        \\
Creatinine Urine     & INR                   & Norepinephrine           & Urine               \\
D5W                  & Insulin Humalog       & Normal Saline             & Vancomycin           \\
DBP                 & Insulin NPH           & O2 Saturation             & Vasopressin          \\
Dextrose Other       & Insulin Regular       & OR/PACU Crystalloid       & WBC                  \\
Dobutamine          & Insulin glargine       & PCO2                      & Weight               \\
Dopamine            & Intubated             & PO intake                 & pH Blood             \\
EBL                  & Jackson-Pratt         & PO2                      & pH Urine             \\
Emesis               & KCl                   & PT                        &                      \\
Eosinophils         & KCl (Bolus)           & PTT                       &          \\
\bottomrule
\end{tabular}
\end{table}

\clearpage 

\section{eICU features}
\label{app:eICU-features}

\begin{table}[!h]
\centering
\caption{For eICU, 100 variables were extracted. The 35 variables on the right column are drug-related. Some of them seem redundant due to different hospitals but can not be merged because of different or not standardized concentrations.}
\label{tab:features_eicu}

\begin{tabular}{ll|l}
\toprule
ALP                & Lactate         & Amiodarone           \\
ALT                & Lymphocytes     & Dobutamine dose      \\
AST                & MBP             & Dobutamine ratio     \\
Albumin            & MCH             & Dopamine dose        \\
Anion Gap          & MCHC            & Dopamine ratio       \\
BUN                & MCV             & Epinephrine dose     \\
Base Deficit       & MPV             & Epinephrine ratio    \\
Base Excess        & Magnesium       & Fentanyl 1           \\
Basophils          & Monocytes       & Fentanyl 2           \\
Bedside Glucose    & Neutrophils     & Fentanyl 3           \\
Bicarbonate        & O2 L/\%         & Furosemide           \\
Bilirubin (Direct) & O2 Saturation   & Heparin 1            \\
Bilirubin (Total)  & PT              & Heparin 2            \\
Bodyweight (kg)    & PTT             & Heparin 3            \\
CO2 (Total)        & PaCO2           & Heparin vol          \\
Calcium            & PaO2            & Insulin 1            \\
Chloride           & Phosphate       & Insulin 2            \\
Creatinine (Blood) & Platelets       & Insulin 3            \\
Creatinine (Urine) & Potassium       & Midazolam 1          \\
DBP                & Protein (Total) & Midazolam 2          \\
Eosinophils       & RBC             & Milrinone 1          \\
EtCO2              & RDW             & Milrinone 2          \\
FiO2               & RR              & Nitroglycerin 1      \\
Fibrinogen         & SBP             & Nitroglycerin 2      \\
GCS eye            & Sodium          & Nitroprusside        \\
GCS motor          & Stool           & Norepinephrine 1     \\
GCS verbal         & Temperature     & Norepinephrine 2     \\
Glucose            & Troponin - I    & Norepinephrine ratio \\
HR                 & Urine           & Pantoprazole         \\
Hct                & WBC             & Propofol 1           \\
Hgb                & pH              & Propofol 2           \\
INR                &                 & Propofol 3           \\
                   &                 & Vasopressin 1        \\
                   &                 & Vasopressin 2        \\
                   &                 & Vasopressin 3              \\
\bottomrule
\end{tabular}
\end{table}

\newpage 

\section{Privacy-Utility Tradeoff Visualization}
\label{app:visu}
\begin{figure}[h!]
    \centering
    \includegraphics[height=0.35\columnwidth, trim=0 0 120 40, clip]{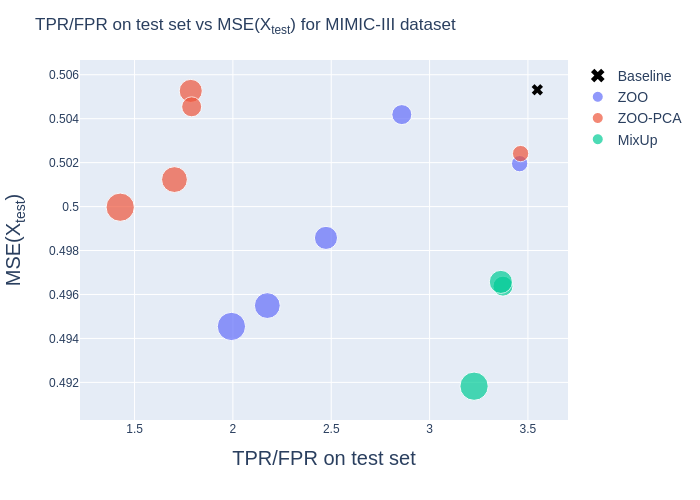}
    \includegraphics[height=0.35\columnwidth, trim=0 0 0 40, clip]{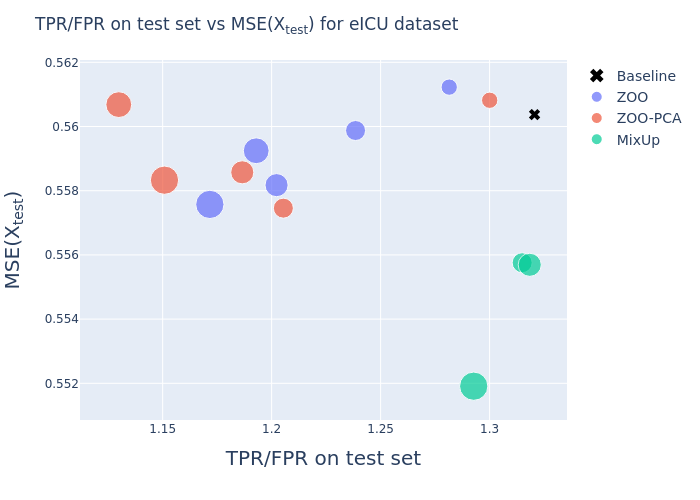} 
    \caption{TPR/FPR ratio of MIAs against generalization performance on test data on MIMIC-III (top) resp. eICU (bottom). The size of the interpolation parameter $\alpha \in \{0, \frac14, \frac12, \frac34, 1\}$ for ZOO and ZOO-PCA, resp. $\beta \in \{0.2,1,5\}$ for MixUp in data augmentation is indicated by the size of the ball. }
    \label{fig:results_plot}
\end{figure}

Figure~\ref{fig:results_plot} plots the TPR/FPR ratio of the loss-based MIA against generalization performance on test data on MIMIC-III. The focus on improvements in mMSE or on privacy protection is shown by the size of the interpolation parameter $\alpha \in \{0, \frac14, \frac12, \frac34, 1\}$. Best results are achieved with higher values of $\alpha$. 

\end{document}